\pdfoutput=1

\documentclass[11pt]{article}
\usepackage{ACL2023}
\usepackage{amsmath}
\usepackage{times}
\usepackage{helvet}
\usepackage{courier}
\usepackage{txfonts}

\usepackage{natbib}  
\usepackage{caption} 

\usepackage{pythonhighlight}
\usepackage{soul}
\usepackage{graphicx}
\usepackage{xcolor}
\usepackage{tabulary}
\usepackage{multirow}
\usepackage{booktabs}
\usepackage{color}
\usepackage{tcolorbox}
\usepackage{newtxtext}
\usepackage{subfigure}
\usepackage{xspace}
\usepackage{hyperref}
\usepackage{listings}

\newcommand{\ie}{\emph{i.e.,}\xspace}
\newcommand{\eg}{\emph{e.g.,}\xspace}

\newcommand{\etc}{\emph{etc.}\xspace}

\tcbset{width=0.5\linewidth,boxrule=0pt,colback=red,arc=0pt,auto outer arc,left=0pt,right=0pt,boxsep=5pt}

\newcommand{\model}{\texttt{K\'EPLET}}
\newcommand{\beyonce}{Beyonc\'e}

\newcommand{\squishlist}{
	\begin{list}{$\bullet$}
		{ \setlength{\itemsep}{0pt}
			\setlength{\parsep}{3pt}
			\setlength{\topsep}{3pt}
			\setlength{\partopsep}{0pt}
			\setlength{\leftmargin}{1.5em}
			\setlength{\labelwidth}{1em}
			\setlength{\labelsep}{0.5em} } }

\newcommand{\squishlisttwo}{
	\begin{list}{$\bullet$}
		{ \setlength{\itemsep}{0pt}
			\setlength{\parsep}{0pt}
			\setlength{\topsep}{0pt}
			\setlength{\partopsep}{0pt}
			\setlength{\leftmargin}{2em}
			\setlength{\labelwidth}{1.5em}
			\setlength{\labelsep}{0.5em} } }
		
\newcommand{\squishend}{
	\end{list}  }

\begin{document}
%
\title{\model: Knowledge-Enhanced Pretrained Language Model \\with Topic Entity Awareness }

\author{Yichuan Li \\ Worcester Polytechnic Institute \\  yli29@wpi.edu 
         \And {\bf Jialong Han} \\ Airbnb \\ jialonghan@gmail.com 
        \And {\bf Kyumin Lee} \\ Worcester Polytechnic Institute \\ kmlee@wpi.edu
        \AND Chengyuan Ma \and Benjamin Yao \and  Derek Liu \\ Amazon Alexa AI \\ chengyuan.ma@gmail.com, benjamy@amazon.com, derecliu@amazon.com
        }

\maketitle
\begin{abstract}

In recent years, Pre-trained Language Models (PLMs) have shown their superiority by pre-training on unstructured text corpus and then fine-tuning on downstream tasks.
On entity-rich textual resources like Wikipedia, Knowledge-Enhanced PLMs (KEPLMs) incorporate the interactions between tokens and \emph{mentioned entities} in pre-training, and are thus more effective on entity-centric tasks such as entity linking and relation classification. 
Although exploiting Wikipedia's rich structures to some extent, conventional KEPLMs still neglect a unique layout of the corpus where each Wikipedia page is around a topic entity (identified by the page URL and shown in the page title).
In this paper, we demonstrate that KEPLMs without incorporating the topic entities will lead to \textit{insufficient entity interaction} and \textit{biased (relation) word semantics}.
We thus propose {\model}, a novel \textbf{\texttt{K}}nowledge-\textbf{\texttt{\'E}}nhanced \textbf{\texttt{P}}re-trained \textbf{\texttt{L}}anguag\textbf{\texttt{E}} model with \textbf{\texttt{T}}opic entity awareness.
In an end-to-end manner, {\model} identifies where to add the topic entity's information in a Wikipedia sentence, fuses such information into token and mentioned entities representations, and supervises the network learning, through which it takes topic entities back into consideration.
Experiments demonstrated the generality and superiority of {\model} which was applied to two representative KEPLMs, achieving significant improvements on four entity-centric tasks.

\end{abstract}

\section{Introduction}

\begin{figure}[t!]
    \centering
    \includegraphics[width=\linewidth]{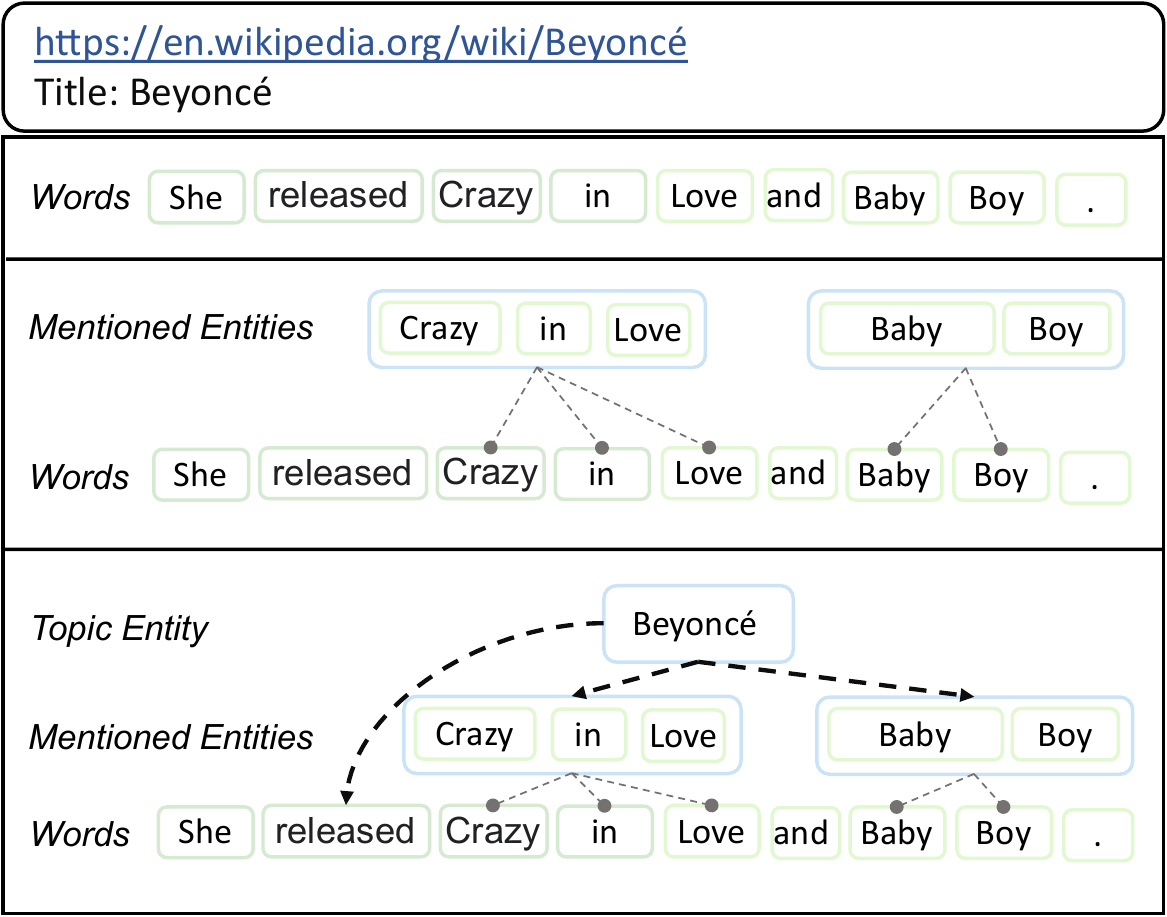}
    \caption{An illustration of Wikipedia page \href{https://en.Wikipedia.org/wiki/Beyonc\%C3\%A9}{{\beyonce}}. There are three levels of interactions among words, mentioned entities and topic entity. We use $\multimapdot$ to stand for the entity and words linkage, and $\rightarrow$ to express an entity interaction need to be considered and word semantics need to be covered by modeling the \emph{topic entity}. 
    }
    \label{fig:Wikipedia_corpus}
\end{figure}

\emph{Pre-trained language models} (PLMs)~\cite{radford2018improving,bert,roberta} have shown their effectiveness on many natural language understanding tasks. 
To exploit the rich syntactic and semantic information in the pre-training data, PLMs are designed to model the word co-occurrences as shown at the top of  \autoref{fig:Wikipedia_corpus}. 
However, they usually fall short in discovering factual knowledge~\cite{lm-factual} and applying such knowledge in language understanding~\cite{zhang2019ernie}.
For example, in sentence \emph{``She released Crazy in Love''} on the Wikipedia page of \textsf{\beyonce}, PLMs will try to mask and predict words like \emph{``Crazy''} and \emph{``in''}, not knowing that \textsf{Crazy in Love} is a \emph{mentioned entity} but tearing it apart.
To incorporate entity knowledge into PLMs, \emph{knowledge-enhanced} PLMs (KEPLMs;~\citeauthor{zhang2019ernie,luke,ERICA,K_BERT}) are proposed to work on not only word co-occurrences but also interactions between words and mentioned entities as well as among the latter. 
As shown in the middle of  \autoref{fig:Wikipedia_corpus}, most KEPLMs work on entity-rich textual resources like Wikipedia, and 
consider hyperlinks of a Wikipedia page as mentions of the target pages' entities to model the rich interactions.
Auxiliary objective functions are also designed to enforce the model to learn entity knowledge in those interactions.
In general, KEPLMs have achieved superior performance compared with PLMs in entity-centric tasks~\cite{survey_rel_word_know,survey_know_intensive}, like named entity recognition, entity typing \etc

KEPLMs have gained their efficacy by modeling entities mentioned in Wikipedia-like textual corpus.
Unfortunately, they still did not fully exploit the entity knowledge there, in the sense that they neglected another important role entities named, \emph{topic entities}, 
which are page titles in the corpus.
For example, in \autoref{fig:Wikipedia_corpus}, those sentences are actually from the page of \textsf{\beyonce} and around this celebrity as a topic.
However, conventional KEPLMs simply neglect such linkages, and treat those sentences independently, which will lead to both \emph{insufficient entity interaction} and \emph{biased (relation) word semantics}.
Take the sentence \emph{``She released \textsf{Crazy in Love} and \textsf{Baby Boy}.''} for example, if ignoring its topic entity \textsf{\beyonce}, KEPLMs can no longer rely on this sentence to capture the interaction between \textsf{Crazy in Love} and \textsf{\beyonce}.
Moreover, the semantics of word ``released'' will be biased since it is between the above pair of entities, not between \textsf{Crazy in Love} and a common word ``she''.

Based on the above discussions, topic entities are indeed important to sentence semantics in Wikipedia-like textual corpus.
However, they are non-trivial to model with simple revisions to existing KEPLMs. 
Readers may wonder whether they can be treated similarly as mentioned entities, \eg assign them certain position embeddings. 
However, we note that it is impractical since most topic entities do not explicitly appear in the sentences like mentioned entities\footnote{In our initial analysis, for top-500K popular Wikipedia pages, only 6\% of these pages mention the topic entity.}.
Readers may also think of using a co-reference resolution model to replace words like ``she'' with the topic model's mention before feeding them to KEPLMs. 
However, embedding such a model to KEPLMs will not only introduce resolution noise, but also be insufficient to cover cases where topic entity information can clarify the local semantics of non-pronoun positions, \eg entity mentions with ambiguous names\footnote{\textsf{``Crazy in Love''} can refer to not only \textsf{\beyonce}'s song, but also an album of \textsf{Itzy}, a Korean girl group.}.


In this paper, we seek to fix the systematic neglect of topic entities in existing KEPLM efforts.
We develop {\model}, a general \textbf{\texttt{K}}nowledge-\textbf{\texttt{\'E}}nhanced \textbf{\texttt{P}}re-trained \textbf{\texttt{L}}anguag\textbf{\texttt{E}} model with \textbf{\texttt{T}}opic entity awareness, which applies to most KEPLMs. 
To exploit mature KEPLMs as its base while filling in the gap that topic entities do not have explicit positions, {\model} features a topic entity fusion module. 
To integrate topic entities into KEPLMs, {\model} first identifies potential \emph{fusing positions} in sentences where topic entities can clarify local ambiguity including but not limited to co-references, through a gated neural network function~\cite{lstm}. 
It then \emph{fuse} the topic entity features into the hidden representations of those positions in a soft manner. 
Finally, {\model} trains the fusion module with the base KEPLM in an end-to-end manner, through a specially designed topic-entity-aware \emph{contrastive loss}.
To validate the generality and effectiveness of {\model}, we conducted comprehensive experiments based on two representative KEPLMs, \ie \texttt{LUKE}~\cite{luke} and \texttt{ERNIE}~\cite{zhang2019ernie}, on entity-centric benchmarks. 
The results demonstrate that {\model} consistently improves the performance of these two KEPLMs across the tasks.


We summarize our contributions as follows: \textbf{1)} We identify the systematic neglect of topic entities in existing KEPLM efforts.
    \textbf{2)} We propose {\model} with a novel topic entity fusion module and a topic-entity-aware loss, forcing existing KEPLMs to fully exploit entity knowledge in Wikipedia-like corpus.
    \textbf{3)} \model~achieves the best performance among existing KEPLMs on several entity-centric benchmarks.

\section{Preliminaries}
\label{sec:background}
In this section, we define two types of entities essential in Wikipedia-based KEPLM pre-training, \ie \emph{mentioned entities} and \emph{topic entities}. 
We also give an overview of how conventional KEPLMs enhance PLMs by incorporating mentioned entities, but neglect topic entities. 

\subsection{ Notations and Definitions}

\noindent\textbf{Mentioned Entities.} On a Wikipedia page, each sentence $S$ consists of a sequence of \emph{tokens} $W=\{w_i\}$ and hyperlinks under some tokens, linking to other Wikipedia pages and forming \emph{mentioned entities}. 
For example, in \autoref{fig:Wikipedia_corpus}, the last sentence with tokens $W=[\emph{She, released, Crazy, in, Love}]$ mentions a song entity \textsf{Crazy in Love} with the last three words. 
In this paper, we denote mentioned entities in a sentence by $E=\{e_i\}$, where each $e_i$ has information about both the entity (\eg Wikipedia URL of \textsf{Crazy in Love}) and the position of the mention span in the sentence. 

\noindent\textbf{Topic Entities.} Besides mentioned entities, every Wikipedia sentence $S$ is also associated with another important entity, the topic entity (denoted by $e_t$), which is the entity of the page where the sentence is from. 
Topic entities are usually identified by the page URLs and indicated by the pages titles.
For example, for the aforementioned sentence in \autoref{fig:Wikipedia_corpus}, the topic entity $e_t$ is \textsf{\beyonce}. 
Although not necessarily mentioning the topic entity $e_t$, all sentences on $e_t$'s page are usually around the topic of discussing all aspects of $e_t$.



\subsection{PLMs, KEPLMs, and Neglect of Topic Entities}

Based on the above notations, every Wikipedia sentence $S$ is essentially a tuple $S=\langle e_t, W, E\rangle$.
On large-scale textual corpus, PLMs~\cite{radford2018improving,bert,roberta}  work on token sequences $W$ by training \emph{transformer}-style encoders via specially designed losses $\mathcal{L}_{PLM}$ (\eg \emph{masked language modeling}).

\noindent\textbf{KEPLMs.}
On top of PLMs, KEPLMs like \texttt{LUKE}~\cite{luke} and \texttt{ERNIE}~\cite{zhang2019ernie} make a further step by utilizing the mentioned entity information in $E$~\cite{survey_rel_word_know}.
Specifically, they extend the encoders to also generate contextualized vectors for the entity mentions, and design entity disambiguation losses $\mathcal{L}_{ED}$ to make those vectors capable of predicating the topic entities. 
To avoid undermining the lexical and synthetic information of the PLMs, KEPLMs pre-train by jointly optimizing its loss and the conventional PLM losses, \ie

\begin{equation}\label{eq:keplm_loss}
    \mathcal{L}_{KEPLM} = \mathcal{L}_{PLM} + \mathcal{L}_{Aux.} \text{.}
\end{equation}

From a holistic point of view, KEPLMs aim to use $\mathcal{L}_{KEPLM}$ to pre-train a language model $\mathcal{LM}$ that can infer contextualized hidden representations for both tokens $W$ and mentioned entities $E$ in a Wikipedia-like sentence, \ie
\begin{align}
    \mathcal{LM}:\langle e_t, W, E\rangle & \rightarrow \langle \mathbf{H}_w, \mathbf{H}_e\rangle \text{,} \\
    \label{eq:keplm}
    \mathcal{LM}:\langle  W, E\rangle & \rightarrow \langle \mathbf{H}_w, \mathbf{H}_e\rangle \text{.}
\end{align}
and fine-tune $\mathcal{LM}$ in entity-centric downstream tasks. 
Note that there is a special hidden vector $\mathbf{h}_{[CLS]}$ in $\mathbf{H}_w$, which is the conventional sentence representation and will be useful in various downstream tasks as well as this work. 


\noindent\textbf{Neglect of Topic Entities.} Although more effective than PLMs in entity-centric tasks, conventional KEPLMs did not fully exploit the rich structure inside a Wikipedia corpora. 
By working on only $W$ and $E$ of a Wikipedia sentence $S=\langle e_t, W, E\rangle$, those KEPLMs neglect the linkage between $S$ and its topic entity $e_t$, leading to two \emph{weaknesses} as follows. 
\squishlist
    \item Besides transformer parameters and initial word vectors, KEPLMs also have a third type of parameters, \ie initial entity vectors, to keep information of entities seen on the pre-training corpora for both topic and mentioned entities. 
    Neglecting $e_t$ of every sentence will deteriorate the learning of vectors for entities occurring more as topic entities but mentioned less.
    \item In a Wikipedia corpus, it is crucial for words representing relations, \eg ``released'' in \emph{``She released `Crazy in Love'\thinspace''}, to learn a good initial representation for downstream tasks like relation classification. 
    Neglecting $e_t$ of every sentence will cause relation words in $W$ to interact only with $E$ rather than both, losing their semantics that they characterize certain relations between $e_t$ and $E$.
\squishend

In this paper, we improve KEPLMs by bringing such linkages back into the architecture and training of $\mathcal{LM}$ in \autoref{eq:keplm}, through which we make up for the above two weaknesses. 


\begin{figure*}[th!]
    \centering
    \includegraphics[width=\linewidth]{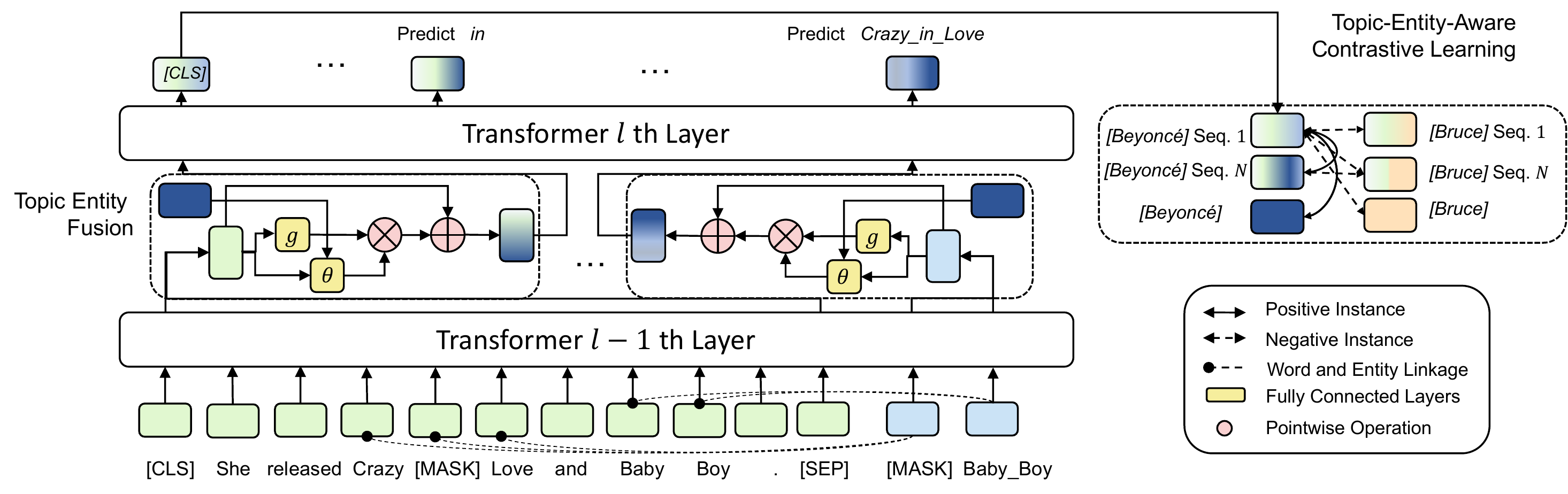}
    \caption{Illustration of {\model} with input \emph{``She released Crazy in Love and Baby Boy''}. The mentioned entities \textsf{Crazy in Love} and \textsf{Baby Boy} are linked with the mentioning words. The middle left part is topic entity fusion module interleaved between transformer layers. The model is trained to predict the masked words and entities as in the top left.Topic-entity-aware contrastive learning is used to enforce the topic entity information by minimizing the distance among sentences' representations under \textsf{{\beyonce}} and hidden representations of \textsf{\beyonce} while enlarging the distance from sentences of other topic entities, \eg \textsf{Bruce}. }
    \label{fig:main_model}
\end{figure*}

\section{\model: Integrating Topic Entities in KEPLMs}


So far, we have motivated the fusion of topic entities into KEPLMs. 
The task then boils down to finding proper places in KEPLM training to fuse \emph{topic entities} information. 
For mentioned entities, existing KEPLMs~\cite{luke,zhang2019ernie,K_BERT,KELM,WKLM,eae} fuse their information by considering them as special tokens spanning multiple positions in a sentence. 
However, topic entities are non-trivial to incorporate in a similar manner, as they do not directly appear in the sentence.
Naively assigning them artificial positions and doing insertions/replacements will potentially break the semantics of the sentences, thus deteriorating the LM's training. 

In this section, we detail how {\model} integrates topic entities.
{\model} features two modules, \emph{fusing position identification} and \emph{entity feature fusion}, and we customize the transformer encoders to accommodate the two modules.
Besides, it employs a topic-entity-aware \emph{contrastive learning} loss.
\autoref{fig:main_model} illustrates the architecture of {\model}.

\subsection{Fusing Position Identification}
\label{sec:position_integrated}


In conventional KEPLMs, the encoders consist of multiple layers of transformers, each of which depends on the contextualized hidden representations of words and mentioned entities from the previous layer. 
Let $\mathbf{H}_S=\text{CONCAT}(\mathbf{H}_w; \mathbf{H}_e)$ be all hidden representations for a Wikipedia sentence output by layer $(l-1)$, and let $\mathbf{h}^{(i)}$ be the $i$-th column of $\mathbf{H}_S$.
From the perspective of preserving the semantics of topic entities $e_t$, we want to find proper positions $i$ to alter the hidden vectors $\mathbf{h}^{(i)}$ with information of $e_t$ before they are sent to the next layer, so that $\mathbf{H}_S$ still carries $e_t$'s semantics even if it is used without knowing what $e_t$ is.

One ideal type of such positions would be pronoun co-references referring to the topic entity, \eg word \emph{``She''} in \emph{``She released `Crazy in Love'\thinspace''}.
They are reasonable places to fuse $e_t$'s information, since the ambiguous semantics of those positions will be clarified after the fusion.
However, we note that co-references are not the only instances of such positions. 
For example,  as word mentions, \textsf{``Crazy in Love''} can refer to not only \textsf{\beyonce}'s song, but also an album of \textsf{Itzy}, a Korean girl group. 
Fusing \textsf{\beyonce}'s information to the three words will thus benefit the representation of this ambiguous span.
To this end, we resort to an end to end approach of identifying fusing positions instead to embed a co-reference resolution model~\cite{corefbert} in {\model}, for the latter cannot cover all cases and is also error-prone due to potentially imperfect model. 
For each position $i$, we compute ${g}_p^{(i)}$ indicating the necessity of fusing $e_t$ to position $i$, \ie
\begin{equation}
    {g}_p^{(i)} = 
    \sigma(F_{p}(\mathbf{h}^{(i)}))\text{.}
    \label{eq:postion_identification}
\end{equation}
Here $F_{p}$ is a \emph{fully connected layer} and $\sigma$ is the sigmoid activation function. To qualitatively evaluate the effectiveness of this module, we did a case study in~\autoref{sec:case_study}.

\subsection{Entity Feature Fusion}\label{sec:method-topic-fusion}

With fusing positions softly identified by ${g}_p^{(i)}$, the entity feature fusion module aims to inject $\textbf{e}_t$, \ie information of the topic entity $e_t$ to the corresponding $\mathbf{h}^{(i)}$.
Inspired by the idea of \emph{Adapters}~\cite{adaptor} interleaved between the transformer layers, we apply an adapter function on $\mathbf{h}^{(i)}$ and $\textbf{e}_t$ to create a fused representation $\hat{\mathbf{h}}^{(i)}$.
We then follow the gating mechanism to softly combine $\hat{\mathbf{h}}^{(i)}$ and the original $\mathbf{h}^{(i)}$ for an updated input $\tilde{\mathbf{{h}}}^{(i)}$ to the next layer $l$, \ie
\begin{align}
    \hat{\mathbf{h}}^{(i)} & = \text{Adapter}(\mathbf{h}^{(i)}, \mathbf{e}_t)\text{,} \\
    \tilde{\mathbf{{h}}}^{(i)} & = \text{LN}((1-{g}_p^{(i)})*  \mathbf{h}^{(i)} + {g}_p^{(i)} * \hat{\mathbf{h}}^{(i)} ) \text{.}
\end{align}
Here $*$ is multiplication and $\text{LN}$ refers to \emph{layer normalization}.


As for $\text{Adapter}(\cdot,\cdot)$, we had two implementations: \emph{concatenation fusion} and \emph{attention fusion}.
We will compare them in the experiment section.

\noindent\textbf{Concatenation Fusion.}  This fusion approach uses a fully connected layer $F_{t}$ to transform the topic entity vector $\mathbf{e}_t$, then concatenates it with the hidden representations $\mathbf{h}^{(i)}$ and feeds the result through another fully connected layer $F_{c}$ as:
\begin{equation}
    \text{Adapter}(\mathbf{h}, \mathbf{e}) = F_{c}(\text{CONCAT}(\mathbf{h};F_{t}(\mathbf{e}_t)))
\end{equation}

\noindent Although the concatenation fusion is quite parameter efficient, it assumes that the words' and mentioned entities' hidden representation $\mathbf{h}$ as well as topic entity vectors $\mathbf{e}_t$ are from an unified feature space. This may be incorrect since the topic entity did not have the position information, while the words and mentioned entities had them.


\noindent\textbf{Attention Fusion.} 
To discern among topic entity, words, and mentioned entities, we propose the attention fusion.
For each hidden vector $\mathbf{h}$, there is an attention fusion around it and the topic entity vector $\mathbf{e}_t$ as follows:
\begin{align}
    \text{Adapter}(\mathbf{h}, \mathbf{e}) & = \text{softmax}\left(\frac{F_q(\mathbf{h})F_k(\mathbf{H})^\top} {\sqrt{d}}\right) F_v(\mathbf{H})  \\
    \mathbf{H} & = \text{CONCAT}(\mathbf{h} ;\mathbf{e}_t)\text{.}
\end{align}
Here $F_q$, $F_k$, $F_v$, are three fully-connected layers to get the query, key and value in the self-attention, and $d$ is the dimension size of hidden vectors $\mathbf{h}_i$.

Finally, we note that all fully connected layers in {\model} are layer-specific, \ie they do not share parameters across different layers in the transformer. 

\subsection{Topic-Entity-Aware Contrastive Loss}
\label{sec:method-obj-pretrain}

To enforce {\model} to really fuse the topic entity information into its output sentence and entity representations, we leverage the co-occurrences of sentences around the same $e_t$ as supervision signals.
We design a novel contrastive learning loss $\mathcal{L}_{t}$, as shown in the right part of \autoref{fig:main_model}. 
The positive pairs are sentences' and topic entity's representation from the same topic entity, while the negative pairs are the sentences' and topic entities' representation from other topic entities in the minibatch.


Concretely, given output sentence representation $\mathbf{h}_{[CLS]}$ from token representations $\mathbf{H}_w$ of a Wikipedia sentence $S$ and its topic entity's representation $\mathbf{e}_t$, we compute the per-sentence loss $\mathcal{L}_S$ as follows. 
We denote $\Delta_S=\{\mathbf{h}_{[CLS]}, \mathbf{e}_t\}$, and all such vectors of Wikipedia sentences on $t$ by 
\begin{equation}
\Delta_t=\bigcup_{S \text{ is on } t}\Delta_S \text{.}
\end{equation}
For each $\mathbf{h}\in \Delta_S$, we draw a positive sample $\mathbf{h}^+$ from $\Delta_t$.
We also draw negative samples $\Delta'$ from vectors $\Delta_{t'}$ for a different topic entity $t'$.
We then compute the contrastive learning loss as follow,
\begin{equation}
    \mathcal{L}_{S} = -\sum_{\mathbf{h}\in \Delta_S}\sum_{\mathbf{h^{+}}\in \Delta_t}{\log\frac{e^{\text{sim}(\mathbf{h}, \mathbf{h^{+}})/\tau}}{\sum_{\mathbf{h}'\in \{\mathbf{h^{+}}\}\cup \Delta'}{e^{\text{sim}(\mathbf{h}, \mathbf{h}')/\tau}}}} \text{.}
\end{equation}
Here ${sim}$ is the cosine similarity between vectors and $\tau$ is a temperature hyperparameter. 
Finally, the overall objective function of {\model} is the sum of $\mathcal{L}_S$ across the corpora and $\mathcal{L}_{KEPLM}$ in \autoref{eq:keplm_loss}.

\section{Experiment Settings}
We evaluate the effectiveness of {\model} on extensive entity-centric tasks: entity typing, relationship classification, named entity recognition and extractive QA. To ensure a fair comparison, we follow the experiment settings of previous work~\cite{luke,zhang2019ernie,eae}. 

\subsection{Baseline Methods}
\label{sec:baseline}
We compare {\model} with the vanilla PLMs: 1) \texttt{BERT}~\cite{bert}, 2) \texttt{RoBERTa}~\cite{roberta}, and KEPLMs: 
3) \texttt{KEPLER}~\cite{KEPLER}, which utilizes the additional knowledge embedding loss to enhance the factual triplets from the knowledge graph for the PLM;
4) \texttt{K-Adapter}~\cite{k-adapter}, which did not explicitly model the entities and adopted the factual knowledge into the external adapter. 
5) \texttt{ERNIE}~\cite{zhang2019ernie}, which injects the mentioned entities' static embedding from KB  and distinct the mentioned entity from the negative sampled entities; 6) \texttt{LUKE}~\cite{luke}, which contains a separated entity embedding and word embedding, and utilizes the auxiliary masked entity prediction besides the masked token prediction to optimize the model. 

\subsection{Implementation Details}
The pretraining Wikipedia corpus is the same as the LUKE~\cite{luke}\footnote{https://archive.org/download/enwiki-20181220}, and we follow the same data preprocessing steps~\cite{luke} to extract entities from hyperlinks. For each entity, we assign a unique entity ID.
The pre-training of {\model} starts from \texttt{LUKE}~\cite{luke} and \texttt{ERNIE}~\cite{zhang2019ernie}, and is optimized for 4.9K steps (1 epoch). 
The temperature $\tau$ for contrastive learning is set to 0.07. The masking entity rate and masked entity rate are set to 60\%. The optimizer for the pre-training is AdamW and is warmed up for 2.5K steps with a learning rate as 1e-5. 
During the downstream tasks' training and evaluation, the topic entity fusion module will be discarded. This is because there is no topic entity for downstream inputs and the knowledge loss for topic entity, and mentioned entity has already been complemented in the KEPLMs during pre-training. For all these baseline methods, we fine-tune their checkpoints on the same hardware and package settings like ~{\model}. 
This may cause the performance gap between our reproduction and their previously reported results. 
The fine-tuning hyperparameter settings are described in \autoref{sec:finetuning}. 

\begin{table}[tb]
    \centering
    \small
    \small
    \addtolength{\tabcolsep}{-0.4em}
    \begin{tabular}{c|ccccc}
    \toprule
    \multirow{2}{*}{Models} & Open Entity & TACRED & SQuAD {1.1} & CoNLL-{2003} \\
    & F\textsubscript{1} & F\textsubscript{1} & F\textsubscript{1}  & F\textsubscript{1} \\
    \midrule
    -Con. & 76.00 & 70.53 & 91.44 &  93.09 \\
    -Atten.  & \textbf{76.38} & \textbf{70.90} &\textbf{92.19} & \textbf{93.64} \\
    
    \bottomrule
    \end{tabular}
    \caption{{Results of different feature fusion modules on \textsf{LUKE}-base+{\model}. }}
    \label{tab:feature_fusion}
\end{table}

\begin{table*}[tbh]
    \centering
    \small
    \addtolength{\tabcolsep}{-0.4em}
    \setlength\tabcolsep{10pt}
    \begin{tabular}{c|ccc|c|cc|ccc}
    \toprule
        \multirow{2}{*}{Models} & \multicolumn{3}{c|}{Open Entity} & TACRED & \multicolumn{2}{c|}{SQuAD {1.1}} & \multicolumn{3}{c}{CoNLL-{2003}} \\
        & Prec. & Rec. & F\textsubscript{1} & F\textsubscript{1} & EM & F\textsubscript{1} & Prec. & Rec. & F\textsubscript{1} \\
        \midrule
        \texttt{BERT}-base &  76.37 & 70.96 & 73.56 & 66.00 & 80.90 & 88.20 & 91.28 & 87.45 &  89.32 \\ 
        \texttt{ERNIE} & \underline{78.42} & \textbf{72.90} & \underline{75.56} & \underline{67.97} & - & - & - & - & -    \\
        \texttt{ERNIE}+{\model} & \textbf{79.85} & \underline{72.01} & \textbf{75.72} & \textbf{70.18} & - & - & - & - & - \\
        \midrule
        \texttt{RoBERTa}-base  &  \textbf{80.62} & 71.43 & 75.74 & 67.95  & \underline{85.18}& 91.49 & \underline{93.24} & 92.49 &  92.86 \\
        \texttt{KEPLER} & 76.78 & \underline{72.43} & 74.54 & \underline{70.70} & - & - &  - & - & -\\
        \texttt{LUKE}-base & \underline{79.70} & 72.37 & \underline{75.86} & 70.30 & \underline{85.18} & \underline{91.88} & 92.81 & \underline{93.27} &  \underline{93.04}\\
        \texttt{LUKE}-base+{\model} & 78.39 & \textbf{74.47} & \textbf{76.38} & \textbf{70.90}  & \textbf{85.66} & \textbf{92.19}  &\textbf{93.48} & \textbf{93.80} & \textbf{93.64} \\
        \midrule
        \texttt{RoBERTa}-large &  79.36 & 73.90 & 76.53 & 71.73 & 87.92 & 94.16 & 91.23 & 92.89 & 92.37 \\
        K-Adapter & 79.30 & \textbf{75.84} & \underline{77.53} & \underline{71.89} & - & - & - & - & - \\
        \texttt{LUKE}-large  &  \underline{79.53} & 75.11 & 77.26 & 71.81 & \underline{88.17} & \underline{94.29} & \textbf{94.27} & \textbf{94.39} &  \textbf{94.33}\\
        \texttt{LUKE}-large+{\model} & \textbf{79.98} & \underline{75.32} & \textbf{77.58} & \textbf{72.33} & \textbf{88.41} & \textbf{94.47} & \underline{94.03} & \underline{94.10} & \underline{94.07} \\
        \bottomrule
    \end{tabular}
    \caption{{Results on entity-centric tasks, \ie entity typing (Open Entity), relationship extraction (TACRED), extractive question answering (SQuAD1.1), and named entity recognition (CoNLL-2003). Bold text indicates best performance and underlined text indicates the second-best performance.
    We mark ``-'' where the corresponding baseline was not previously applied on this task. 
    }}
    \label{tab:exp_result_entity_centric}
\end{table*}

\subsection{Entity-Centric Tasks}
\label{sec:exp_tasks}
The fine-tuning of entity centric tasks cannot simply take the hidden representation of ``[CLS]'' to represent the whole sentence. These tasks require special procedures to better represent the entities inside the sentence. Readers can refer to ~\cite{luke,zhang2019ernie} for detailed fine-tuning procedures.
The summaries and evaluation metrics for these entity centric tasks are as follows:

\noindent\textbf{Entity Typing}
is to predict the types of an entity given the entity mention and the contextual sentence around the entity mention. Following the previous experiment setting of~\cite{zhang2019ernie,luke}, we use the Open Entity dataset~\cite{choi-etal-2018-ultra} and only consider nine popular entity types. We report the precision, recall and micro-F\textsubscript{1} scores, and use the micro-F\textsubscript{1} score for comparison.

\noindent\textbf{Relation Classification}
is to classify the correlation between two entities. The input data includes two entity mentions and contextual information around these two entities. We utilize TACRED dataset~\cite{zhang-etal-2017-position} and also report the micro-F\textsubscript{1} score for the comparison.

\noindent\textbf{Named Entity Recognition}
is to identify the entity from the given sentence. We utilize CoNLL-2003~\cite{conll2003} dataset and report the span-level F\textsubscript{1} scores. 

\noindent\textbf{Extractive Question Answering} is to answer a question by extracting text span from a given passage. We utilize SQuAD1.1~\cite{squad1} and report the exact match~(EM) and token-level F\textsubscript{1} on the development dataset. 

\section{Experiment Results}
In this section, we aim to answer the following experimental questions: \textbf{EQ1} Which feature fusion method can better integrate the topic entity with KEPLMs? \textbf{EQ2} Can topic entity fusion improve the performance of KEPLMs on downstream tasks? 
and \textbf{EQ3} How should the topic entity fusion module of \texttt{\model} be added to KEPLMs to achieve optimal performance?

\subsection{Comparing Entity Feature Fusion Approaches}\label{sec:exp_adapter}

To answer \textbf{EQ1}, we conduct the experiment on entity typing and relation extraction tasks and make comparison between {\model}-Con and {\model}-Atten.
As the results are shown in  \autoref{tab:feature_fusion}, we can observe that {\model}-Atten achieves better performance than {\model}-Con in all the tasks. This indicates there may exist the unnecessary information or noise from the topic entity towards the words and mentioned entities, so it is important to selectively to do the feature fusion. Based on this observation, in the following sections, we will only report better feature fusion model: {\model}-Atten. 

\subsection{Effectiveness of Topic Entity Fusion}
To answer \textbf{EQ2}, we compare the performance of {\model} among the PLMs and KEPLMs.
We report the experiment results in \autoref{tab:exp_result_entity_centric}. 
Note that in the literature, different KEPLMs may be initialized with different base PLMs, \eg \texttt{ERNIE} with \texttt{BERT}-base but \texttt{KEPLER} with \texttt{RoBERTa}-base.
Therefore, we conduct three groups of comparisons, each corresponding one of the three possible base PLMs (\ie, \texttt{BERT}-base, \texttt{RoBERTa}-base, and \texttt{RoBERTa}-large) as well as the baseline KEPLMs basing on it. 
\texttt{\model} is implemented on the best performing baseline KEPLMs, \ie \texttt{ERNIE} or \texttt{LUKE}, whose performance is reported in the same group. 
In general, we have the following observations:

\noindent \textbf{Effectiveness of KEPLMs in entity-centric tasks.} We observe that in most cases KEPLMs achieve better performance compared with their initialized language models. This is because KEPLMs can better capture the entity-level factual knowledge besides the syntax based word co-occurrence information. 

 \noindent \textbf{Effectiveness of advanced KEPLMs.} We observe that \texttt{LUKE} achieves better performance than \texttt{ERNIE} on all the entity-centric tasks. This is because the \texttt{LUKE} can learn the better contextual embedding of mentioned entities, while the \texttt{ERNIE} utilizes the static entity embedding.

\noindent\textbf{Effectiveness of {\model} over KEPLMs.} By considering the topic entity knowledge, {\model} gets considerable improvement on most tasks, achieving the best or second-best performance within the same comparison group. Specifically, on TACRED, \texttt{\model} gets {0.6\%} performance improvement compared with \texttt{LUKE}-base\footnote{ Our performance improvement is compatible with other SOTA models like \textsf{LUKE} and \textsf{ERICA} performance improvement. These SOTA methods also make 0.x\% improvements compared with their baseline methods. 
}.
This not only indicates the importance of integrating the topic entity into KEPLMs pretraining, but also represent the generalization of {\model}.   

Overall, {\model} achieves the best performance on all the datasets and metrics.  {\model} brings the consistent performance improvement over the backbone KEPLMs.
\begin{figure}[th!]
    \centering
    \includegraphics[width=\linewidth]{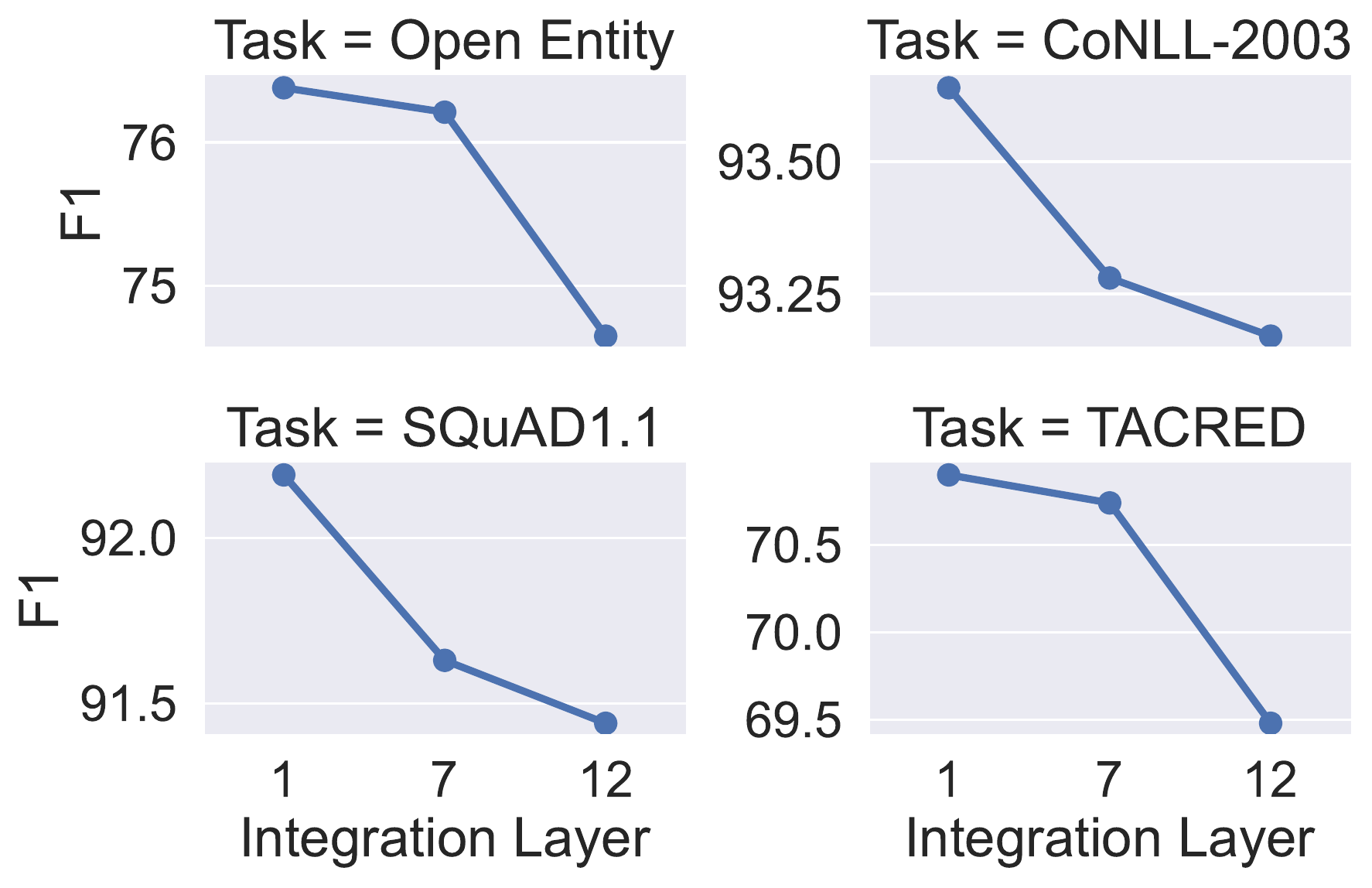}
    \caption{Parameter analysis of topic entity fusion layer $l$ on \texttt{LUKE}-base+{\model}. }
    \label{fig:layer_analsysi}
\end{figure}

\begin{table}[th!]
    \centering
    \small
    \small
    \addtolength{\tabcolsep}{-0.4em}
    \begin{tabular}{c|ccccc}
    \toprule
    \multirow{2}{*}{\# Layers} & Open Entity & TACRED & SQuAD {1.1} & CoNLL-{2003} \\
     &  F\textsubscript{1} & F\textsubscript{1} & F\textsubscript{1}  & F\textsubscript{1} \\
    \midrule
    1 &  \textbf{76.38} & \textbf{70.90} &\textbf{92.19} & \textbf{93.64} \\
    2   &  \underline{76.21} & \underline{70.80} & \underline{91.60} &  {93.12} \\
    3 & 75.66   & 70.01 & 91.58 & \underline{93.31}  \\
    \bottomrule
    \end{tabular}
    \caption{{Results of different number of topic entity fusion module layers. }}
    \label{tab:layer_number}
\end{table}

\subsection{Parameter Analysis}
To address \textbf{EQ3}, we vary where and how many layers of topic entity fusion modules to be added to the \texttt{LUKE}-base model, which has 12 transformer layers. 
We first conduct a parameter analysis of adding the fusion module between layers $l$ and $l-1$" layer $l$.
As shown in \autoref{fig:layer_analsysi}, we can observe that fusion at lower layer achieves better performance than higher layer. Specifically, $l=1$ achieves the best performance compared with $l=7 and 12$. Since the lower layers contain more general information and the higher layers contain more task-specific information~\cite{bert_layer}, the topic entity can bring the general and comprehensive knowledge for the mentioned entities and word semantics.

In addition, we also evaluate the effectiveness of inserting different numbers of topic entity fusion modules in lower layers. We make the comparison among the single, double and triple topic entity fusion modules on \textsf{LUKE-base} backbone. As the results shown in \autoref{tab:layer_number}, single position fusion layer achieves the best performance. This observation represents the efficiency of feature fusion module that can bring enough topic entity knowledge by only doing one time feature fusion.

\section{Related Work} 



We group the related KEPLMs works based on what context information is used when encoding sentences with entity knowledge (\ie mentions), which are depending on \emph{sentential context only} or additional \emph{non-sentential context}.

\noindent\textbf{Sentential Context Only KEPLMs.}
Most KEPLMs pre-trains by encoding sentences with mentioned entities and aligning the encoded spans with their ground truth entities. 
\texttt{ERNIE}~\cite{zhang2019ernie} integrates static entity embeddings into their PLM with a novel fusion layer to fuse spans' representations with entities'.
\texttt{LUKE}~\cite{luke} treats both words and entities as tokens but uses different embeddings, self-attention layers, and masked token/entity prediction heads. 
It averages position embeddings of words in a span as the position embedding of the entity.
Different from \texttt{ERNIE} and \texttt{LUKE}, which require mention spans to be available,
\texttt{KnowBERT}~\cite{K_BERT} and \texttt{EAE}~\cite{eae} incorporate entity linking modules to identify the spans, which are trained in an end-to-end manner.
To deal with entity co-references in sentences, \texttt{CorefBERT}~\cite{corefbert} and \texttt{TOME}~\cite{mention_memory} propose extending the alignments to not only explicit entity mentions but also implicit co-reference spans. 
\texttt{WKLM}~\cite{WKLM} pre-trains by distinguishing the ground truth entity's embeddings from those of randomly corrupted ones. 
\texttt{K-Adaptor}~\cite{k-adapter} utilizes plug-in adapters~\cite{adaptor} to inject factual and linguistic knowledge without updating the LMs. 

\noindent\textbf{Non-Sentential Context KEPLMs.}
While KEPLMs depending only on sentential context have shown efficacy, other works argue that the alignment between mention spans and entities can be more effectively done if the mention spans are enriched with additional context outside the current sentence. 
\texttt{GRAPHCACHE}\cite{graphcache} constructs non-sentential context by building a heterogeneous graph with sentence and property nodes,
and their interaction edges.
Besides textual context, there are also works~\cite{KEPLER,K_BERT,bert_mk,CokeBERT,colake,SMedBERT} utilizing external knowledge graphs (KGs) as more informative non-sentential context. 
For example, \texttt{KEPLER}~\cite{KEPLER} injects mentioned entities' descriptions and KG facts into its PLM.
\texttt{K-BERT}~\cite{K_BERT} fuses KG facts into sentences by constructing a sentence tree.
Compared with textual context, KG-based context often requires specially processed and linked knowledge from Wikipedia or WikiData, where noises could be introduced in those processes. 
In our work, we explore topic entities as a unique non-sentential context that is directly available in the Wikipedia structure without additional processing.

\noindent\textbf{Contrastive Learning in KEPLMs.} Orthogonal to various types of context, multiple objectives or tasks have been introduced by previous studies to pre-train those KEPLMs.
Most of those attempts fall in the contrastive learning setting to minimize distances between positive pairs while enlarging those between negative ones~\cite{contrastive_raw,simcse,simlr,coco_lm}.
Therefore, the construction of positive pairs and negative pairs is crucial in this setting. 
In \texttt{UCTopic}~\cite{uctopic}, positive pairs are sentences that share the same entity mentions, while negative ones are from different mentions. 
\texttt{EASE}~\cite{ease} extends \texttt{UCTopic} by incorporating entity representations into positive pairs.
\texttt{LinkBERT}~\cite{yasunaga2022linkbert} introduces a sentence relation classification objective to make sentence representations capable of predicting sentences' relations. 
Unlike those works, we construct positive and negative pairs with the unique topic entity information to guide the semantic learning of entities and relation words. 

\section{Conclusion}
We propose {\model} to bring back knowledge on topic entities that are systematically neglected in KEPLM learning. 
{\model} performs topic entity fusion by identifying the potential fusing positions and fusing topic entity features through concatenation- or attention-based fusion manners.
In addition, we design a new pre-training task, \ie topic-entity-aware contrastive learning, for better topic entity fusion.
Experiment results on several entity-centric tasks prove the effectiveness of {\model}. 
Potential directions of the future work include: \textbf{1)} Adopting {\model} to multi-language setting.
Since topic entities are language invariant~\cite{ease}, we can expand the topic-entity-aware contrastive learning for the same topic entity under different languages; \textbf{2)} Applying {\model} to domain-specific knowledge-centric tasks like fact checking and fake news detection.

\clearpage

\section{Limitations}
Since our method has incorporated the topic entity into language model pretraining, it requires an entity-rich pretraining dataset. 
The dataset's layout should have topic entity and mentioned entities. 
The demand for special pretraining dataset is a clear limitation of our method.  
In addition, our pretraining stage require many computational resources (eight A10 cards for up to 16 hours, so totally 128 GPU hours). 
This is because we want to inject the previously neglected topic entity knowledge from Wikipedia into KEPLMs. 
This will require us to do the further pretraining on the whole Wikipedia dataset. 
However, given the performance improvements brought by our method on many entity-centric tasks, the additional computational cost is totally worthy. 


\section{Ethical Considerations}
The training data is publicly available and there is no violation of personal privacy. 
To make a fair comparison, we conduct all the experiments under the same hardware and package settings. 
We also report the detailed experiment settings and hyper-parameter to improve the work reproducibility.
In this work, we spent many computation resources for tuning the model structures. 
However, in inference, our model can easily fit into one RTX1080 card. This will improve the democratic values of our model.

\bibliography{cite}
\bibliographystyle{acl_natbib}
\appendix

\section{Details of Pretraining}
Following the work of \textsf{LUKE}\cite{luke}, we utilize the same pretraining dataset: December 2018 version of Wikipedia\footnote{https://github.com/studio-ousia/luke/issues/112} and hyperparameters, except learning rate (1e-6). 
We separate different input sequences from same Wikipedia page by 512 words. 
To provide enough positive pairs for Topic-Entity-Aware Contrastive Loss, we modify the code of data minibatch generation for pretraining. 
Specifically, we firstly combine consecutive elements of the pretraining dataset into batches, then do the shuffling and lastly split the batch into multiple elements. 
The TensorFlow-style code block is like follows:
\begin{python}
 # LUKE's implementation
 data = data.shuffle()
 # Our implementation
data = data.batch(2).shuffle().unbatch()    
\end{python}

\section{Details for Fine-Tuning}
\label{sec:finetuning}
Following the hyper-parameter setting from LUKE~\cite{luke}, we conduct the hyper-parameter search on all the datasets except SQuAD1.1. 
We use grid search to find the best model based on the validation performance. 
The metrics for model selection are reported in \autoref{sec:exp_tasks}.
All in all, we use the following search space:
\squishlist
\item learning rate 1e-5, 2e-5, 5e-5, 5e-6
\item batch size: 4, 8, 16, 32, 64
\item number of training epochs: 2, 3, 5
\squishend


\section{Case Study}
\label{sec:case_study}
We use a case study to illustrate which part of a sentence will be considered by \texttt{\model} as potential integrated positions.
As is shown in \autoref{tab:case_study_inserted_position_pl}, we visualize the integrated positions' rankings, based on the rank of $g_p^{(i)}$, from \textsf{LUKE}-base backbone. 
We can observe that the integrated positions usually involve the pronouns, \ie ``he'' and mentioned entities, \ie \texttt{Farrukhsiyar} and \texttt{Western text-type} \etc. 
This is consistent with the idea case that the topic entity can improve the syntax of the sentence and enrich the representation of the mentioned entities. This observation validates that {\model} can bring the general topic entity knowledge instead of the task-specific information. 
\begin{table*}[tbh!]
\small
    \centering
    \begin{tabular}{p{0.145\linewidth}|p{0.8\linewidth}}
    \toprule
    Topic Entity & Words and Mentioned Entities \\
    \midrule
    \href{https://en.Wikipedia.org/wiki/Hugh\_Bonneville}{Hugh Bonneville}
   &  {\setlength{\fboxsep}{0pt}\colorbox{white!0}{\parbox{1\linewidth}{
\colorbox{blue!0.0}{\strut Hugh} \colorbox{blue!43.0}{\strut Williams} \colorbox{blue!0.0}{\strut Jared} \colorbox{blue!0.0}{\strut November} \colorbox{blue!0.0}{\strut ACP} \colorbox{blue!0.0}{\strut ),} \colorbox{blue!3.0}{\strut professionally} \colorbox{blue!0.0}{\strut as} \colorbox{blue!0.0}{\strut Bon} \colorbox{blue!0.0}{\strut ne} \colorbox{blue!0.0}{\strut ville} \colorbox{blue!0.0}{\strut is} \colorbox{blue!0.0}{\strut English} \colorbox{blue!0.0}{\strut crew} \colorbox{blue!19.0}{\strut is} \colorbox{blue!0.0}{\strut best} \colorbox{blue!0.0}{\strut 408} \colorbox{blue!18.0}{\strut Robert} \colorbox{blue!0.0}{\strut Craw} \colorbox{blue!0.0}{\strut ley} \colorbox{blue!0.0}{\strut in} \colorbox{blue!0.0}{\strut the} \colorbox{blue!0.0}{\strut series} \colorbox{blue!0.0}{\strut CRA} \colorbox{blue!0.0}{\strut (} \colorbox{blue!0.0}{\strut 2010} \colorbox{blue!0.0}{\strut 2015} \colorbox{blue!0.0}{\strut surging} \colorbox{blue!0.0}{\strut he} \colorbox{blue!0.0}{\strut a} \colorbox{blue!0.0}{\strut Globe} \colorbox{blue!12.0}{\strut Award} \colorbox{blue!4.0}{\strut two} \colorbox{blue!41.0}{\strut Emmy} \colorbox{blue!0.0}{\strut rockets} \colorbox{blue!1.0}{\strut SHARE} \colorbox{blue!0.0}{\strut .} \colorbox{blue!0.0}{\strut was} \colorbox{blue!35.0}{\strut born} \colorbox{blue!0.0}{\strut in} \colorbox{blue!0.0}{\strut Tues} \colorbox{blue!0.0}{\strut puck} \colorbox{blue!0.0}{\strut folder} \colorbox{blue!0.0}{\strut Oscar} \colorbox{blue!0.0}{\strut ,} \colorbox{blue!0.0}{\strut to} \colorbox{blue!9.0}{\strut ight} \colorbox{blue!0.0}{\strut who} \colorbox{blue!0.0}{\strut SAY} \colorbox{blue!0.0}{\strut father} \colorbox{blue!0.0}{\strut who} \colorbox{blue!49.0}{\strut a} \colorbox{blue!0.0}{\strut .} \colorbox{blue!27.0}{\strut \textbf{He}} \colorbox{blue!0.0}{\strut was} \colorbox{blue!0.0}{\strut educated} \colorbox{blue!0.0}{\strut ributes} \colorbox{blue!17.0}{\strut clothing} \colorbox{blue!23.0}{\strut College} \colorbox{blue!0.0}{\strut at} \colorbox{blue!0.0}{\strut Sher} \colorbox{blue!0.0}{\strut borne} \colorbox{blue!0.0}{\strut School} \colorbox{blue!10.0}{\strut ,} \colorbox{blue!46.0}{\strut independent} \colorbox{blue!28.000000000000004}{\strut in} \colorbox{blue!0.0}{\strut D} \colorbox{blue!5.0}{\strut orset} \colorbox{blue!16.0}{\strut .} \colorbox{blue!0.0}{\strut Following} \colorbox{blue!0.0}{\strut secondary} \colorbox{blue!0.0}{\strut Bon} \colorbox{blue!0.0}{\strut ne} \colorbox{blue!0.0}{\strut ville} \colorbox{blue!0.0}{\strut at} \colorbox{blue!0.0}{\strut Corpus} \colorbox{blue!0.0}{\strut Christ} \colorbox{blue!6.0}{\strut i} \colorbox{blue!0.0}{\strut ,} \colorbox{blue!0.0}{\strut ,} \colorbox{blue!0.0}{\strut PASS} \colorbox{blue!0.0}{\strut predictions} \colorbox{blue!0.0}{\strut Web} \colorbox{blue!0.0}{\strut ber} \colorbox{blue!0.0}{\strut Douglas} \colorbox{blue!0.0}{\strut of} \colorbox{blue!0.0}{\strut .} \colorbox{blue!0.0}{\strut Dire} \colorbox{blue!25.0}{\strut Cambridge} \colorbox{blue!0.0}{\strut a} \colorbox{blue!28.999999999999996}{\strut exorc} \colorbox{blue!0.0}{\strut beer} \colorbox{blue!36.0}{\strut in} \colorbox{blue!0.0}{\strut theology} \colorbox{blue!0.0}{\strut that} \colorbox{blue!26.0}{\strut \textbf{he}} \colorbox{blue!0.0}{\strut tended} \colorbox{blue!0.0}{\strut to} \colorbox{blue!0.0}{\strut do} \colorbox{blue!47.0}{\strut bath} \colorbox{blue!0.0}{\strut than} \colorbox{blue!44.0}{\strut academic} \colorbox{blue!0.0}{\strut work} \colorbox{blue!0.0}{\strut Bon} \colorbox{blue!0.0}{\strut ne} \colorbox{blue!0.0}{\strut ville} \colorbox{blue!20.0}{\strut is} \colorbox{blue!0.0}{\strut of} \colorbox{blue!0.0}{\strut the} \colorbox{blue!0.0}{\strut Youth} \colorbox{blue!0.0}{\strut Theatre} \colorbox{blue!0.0}{\strut .} \colorbox{blue!0.0}{\strut Bon} \colorbox{blue!21.0}{\strut ne} \colorbox{blue!0.0}{\strut ville} \colorbox{blue!0.0}{\strut 's} \colorbox{blue!0.0}{\strut professional} \colorbox{blue!0.0}{\strut IRC} \colorbox{blue!0.0}{\strut appearance} \colorbox{blue!0.0}{\strut was} \colorbox{blue!0.0}{\strut the} \colorbox{blue!0.0}{\strut Open} \colorbox{blue!0.0}{\strut Air} \colorbox{blue!13.0}{\strut avalanche} \colorbox{blue!33.0}{\strut .} \colorbox{blue!0.0}{\strut In} \colorbox{blue!22.0}{\strut 1987} \colorbox{blue!0.0}{\strut Theatre} \colorbox{blue!0.0}{\strut foreground} \colorbox{blue!0.0}{\strut he} \colorbox{blue!0.0}{\strut the} \colorbox{blue!0.0}{\strut Royal} \colorbox{blue!0.0}{\strut 1991} \colorbox{blue!0.0}{\strut ,} \colorbox{blue!0.0}{\strut where} \colorbox{blue!0.0}{\strut 1966} \colorbox{blue!0.0}{\strut 's} \colorbox{blue!0.0}{\strut Ham} \colorbox{blue!0.0}{\strut let} \colorbox{blue!0.0}{\strut (} \colorbox{blue!0.0}{\strut 1992} \colorbox{blue!0.0}{\strut 1993} \colorbox{blue!0.0}{\strut ).} \colorbox{blue!15.0}{\strut \textbf{He}} \colorbox{blue!0.0}{\strut in} \colorbox{blue!0.0}{\strut railroad} \colorbox{blue!0.0}{\strut of} \colorbox{blue!32.0}{\strut '} \colorbox{blue!0.0}{\strut T} \colorbox{blue!45.0}{\strut is} \colorbox{blue!0.0}{\strut 's} \colorbox{blue!0.0}{\strut a} \colorbox{blue!0.0}{\strut Wh} \colorbox{blue!0.0}{\strut ore} \colorbox{blue!0.0}{\strut ,} \colorbox{blue!0.0}{\strut K} \colorbox{blue!0.0}{\strut ast} \colorbox{blue!0.0}{\strut ril} \colorbox{blue!0.0}{\strut and} \colorbox{blue!0.0}{\strut later} \colorbox{blue!0.0}{\strut in} \colorbox{blue!0.0}{\strut The} \colorbox{blue!14.000000000000002}{\strut Alchemist} \colorbox{blue!0.0}{\strut .} \colorbox{blue!0.0}{\strut ,} \colorbox{blue!0.0}{\strut Bon} \colorbox{blue!0.0}{\strut ne} \colorbox{blue!0.0}{\strut ville} \colorbox{blue!0.0}{\strut made} \colorbox{blue!0.0}{\strut television} \colorbox{blue!0.0}{\strut debut} \colorbox{blue!0.0}{\strut ,} \colorbox{blue!0.0}{\strut as} \colorbox{blue!0.0}{\strut Bon} \colorbox{blue!0.0}{\strut ne} \colorbox{blue!2.0}{\strut ville} \colorbox{blue!0.0}{\strut .} \colorbox{blue!42.0}{\strut \textbf{His}} \colorbox{blue!40.0}{\strut film} \colorbox{blue!0.0}{\strut was} \colorbox{blue!0.0}{\strut 1994} \colorbox{blue!0.0}{\strut 's} \colorbox{blue!0.0}{\strut Mary} \colorbox{blue!0.0}{\strut Shelley} \colorbox{blue!0.0}{\strut opot} \colorbox{blue!0.0}{\strut and} \colorbox{blue!8.0}{\strut Kenneth} \colorbox{blue!0.0}{\strut roles} \colorbox{blue!0.0}{\strut -} \colorbox{blue!0.0}{\strut nat} \colorbox{blue!0.0}{\strut ured} \colorbox{blue!0.0}{\strut b} \colorbox{blue!0.0}{\strut umbling} \colorbox{blue!0.0}{\strut characters} \colorbox{blue!0.0}{\strut like} \colorbox{blue!30.0}{\strut (} \colorbox{blue!0.0}{\strut 1999} \colorbox{blue!0.0}{\strut Mr} \colorbox{blue!0.0}{\strut Park} \colorbox{blue!11.0}{\strut Gap} \colorbox{blue!0.0}{\strut BBC} \colorbox{blue!0.0}{\strut series} \colorbox{blue!0.0}{\strut ,} \colorbox{blue!0.0}{\strut Armored} \colorbox{blue!0.0}{\strut (} \colorbox{blue!0.0}{\strut 2000} \colorbox{blue!0.0}{\strut and} \colorbox{blue!0.0}{\strut played} \colorbox{blue!0.0}{\strut more} \colorbox{blue!0.0}{\strut villain} \colorbox{blue!0.0}{\strut ous} \colorbox{blue!0.0}{\strut characters} \colorbox{blue!0.0}{\strut ,} \colorbox{blue!38.0}{\strut leading} \colorbox{blue!0.0}{\strut the} \colorbox{blue!0.0}{\strut dom} \colorbox{blue!0.0}{\strut ine} \colorbox{blue!0.0}{\strut ering} \colorbox{blue!0.0}{\strut Hen} \colorbox{blue!37.0}{\strut leigh} \colorbox{blue!0.0}{\strut Der} \colorbox{blue!0.0}{\strut onda} \colorbox{blue!0.0}{\strut (} \colorbox{blue!31.0}{\strut 2002} \colorbox{blue!0.0}{\strut )} \colorbox{blue!0.0}{\strut and} \colorbox{blue!0.0}{\strut resilience} \colorbox{blue!0.0}{\strut seized} \colorbox{blue!0.0}{\strut risky} \colorbox{blue!7.000000000000001}{\strut )} \colorbox{blue!0.0}{\strut In} \colorbox{blue!0.0}{\strut Again} \colorbox{blue!0.0}{\strut he} \colorbox{blue!0.0}{\strut played} \colorbox{blue!0.0}{\strut hours} \colorbox{blue!0.0}{\strut poet} \colorbox{blue!0.0}{\strut L} \colorbox{blue!0.0}{\strut arkin} \colorbox{blue!0.0}{\strut .} \colorbox{blue!0.0}{\strut Iris} \colorbox{blue!0.0}{\strut ),} \colorbox{blue!0.0}{\strut young} \colorbox{blue!39.0}{\strut Bay} \colorbox{blue!0.0}{\strut ley} \colorbox{blue!0.0}{\strut Orth} \colorbox{blue!0.0}{\strut his} \colorbox{blue!0.0}{\strut by} \colorbox{blue!0.0}{\strut and} \colorbox{blue!0.0}{\strut B} \colorbox{blue!0.0}{\strut AFTA} \colorbox{blue!0.0}{\strut In} \colorbox{blue!0.0}{\strut Mountain} \colorbox{blue!0.0}{\strut Luther} \colorbox{blue!0.0}{\strut in} \colorbox{blue!0.0}{\strut The} \colorbox{blue!0.0}{\strut Man} \colorbox{blue!0.0}{\strut ridicule} \colorbox{blue!0.0}{\strut .} \colorbox{blue!0.0}{\strut Bon} \colorbox{blue!0.0}{\strut ne} \colorbox{blue!0.0}{\strut ville} \colorbox{blue!0.0}{\strut works} \colorbox{blue!0.0}{\strut in} \colorbox{blue!0.0}{\strut radio} \colorbox{blue!0.0}{\strut MIA} \colorbox{blue!0.0}{\strut \#Joachim Murat\#} \colorbox{blue!24.0}{\strut \textbf{\#Primetime Emmy Award for Outstanding Lead Actor in a Drama Series\#}} \colorbox{blue!0.0}{\strut \#Paddington\#} \colorbox{blue!0.0}{\strut \#National Youth Theatre\#} \colorbox{blue!0.0}{\strut \#Stafford\#} \colorbox{blue!0.0}{\strut \#2014 FIFA World Cup\#} \colorbox{blue!0.0}{\strut \#Single-elimination tournament\#} \colorbox{blue!48.0}{\strut \textbf{\#Gia Long\#}} \colorbox{blue!34.0}{\strut \textbf{\#Farrukhsiyar\#}} \colorbox{blue!0.0}{\strut \#Bridget Jones's Diary\#} \colorbox{blue!0.0}{\strut \#Cousin\#} \colorbox{blue!0.0}{\strut \#Juicy Couture\#} \colorbox{blue!0.0}{\strut \#Nationalism\#} \colorbox{blue!0.0}{\strut \#British Rail Class 08\#} 
}}} \\
    \midrule
    \href{https://en.Wikipedia.org/wiki/Lucius\_Valerius\_Messalla\_Thrasea\_Priscus}{Lucius Valerius Messalla Thrasea Priscus
}
 & {\setlength{\fboxsep}{0pt}\colorbox{white!0}{\parbox{\linewidth}{
\colorbox{blue!46.0}{\strut Val} \colorbox{blue!0.0}{\strut er} \colorbox{blue!0.0}{\strut ius} \colorbox{blue!30.0}{\strut Thr} \colorbox{blue!0.0}{\strut ase} \colorbox{blue!0.0}{\strut a} \colorbox{blue!20.0}{\strut (} \colorbox{blue!9.0}{\strut d} \colorbox{blue!0.0}{\strut ied} \colorbox{blue!5.0}{\strut c} \colorbox{blue!0.0}{\strut .} \colorbox{blue!0.0}{\strut was} \colorbox{blue!0.0}{\strut a} \colorbox{blue!44.0}{\strut Roman} \colorbox{blue!0.0}{\strut active} \colorbox{blue!0.0}{\strut the} \colorbox{blue!0.0}{\strut Grind} \colorbox{blue!0.0}{\strut televised} \colorbox{blue!0.0}{\strut of} \colorbox{blue!7.000000000000001}{\strut Sept} \colorbox{blue!0.0}{\strut imus} \colorbox{blue!0.0}{\strut He} \colorbox{blue!38.0}{\strut was} \colorbox{blue!0.0}{\strut cons} \colorbox{blue!0.0}{\strut ul} \colorbox{blue!0.0}{\strut 196} \colorbox{blue!0.0}{\strut inate} \colorbox{blue!3.0}{\strut colleague} \colorbox{blue!0.0}{\strut G} \colorbox{blue!24.0}{\strut ai} \colorbox{blue!1.0}{\strut us} \colorbox{blue!0.0}{\strut .} \colorbox{blue!0.0}{\strut at} \colorbox{blue!28.000000000000004}{\strut Dating} \colorbox{blue!13.0}{\strut tally} \colorbox{blue!0.0}{\strut was} \colorbox{blue!4.0}{\strut a} \colorbox{blue!33.0}{\strut of} \colorbox{blue!0.0}{\strut g} \colorbox{blue!39.0}{\strut ens} \colorbox{blue!43.0}{\strut .} \colorbox{blue!0.0}{\strut It} \colorbox{blue!0.0}{\strut is} \colorbox{blue!23.0}{\strut V} \colorbox{blue!0.0}{\strut ip} \colorbox{blue!0.0}{\strut stan} \colorbox{blue!0.0}{\strut us} \colorbox{blue!0.0}{\strut ,} \colorbox{blue!2.0}{\strut who} \colorbox{blue!0.0}{\strut may} \colorbox{blue!0.0}{\strut have} \colorbox{blue!0.0}{\strut Statements} \colorbox{blue!0.0}{\strut a} \colorbox{blue!16.0}{\strut design} \colorbox{blue!0.0}{\strut atus} \colorbox{blue!0.0}{\strut but} \colorbox{blue!0.0}{\strut died} \colorbox{blue!49.0}{\strut \textbf{he}} \colorbox{blue!0.0}{\strut to} \colorbox{blue!0.0}{\strut the} \colorbox{blue!0.0}{\strut consulate} \colorbox{blue!0.0}{\strut so} \colorbox{blue!28.999999999999996}{\strut ,} \colorbox{blue!0.0}{\strut Thr} \colorbox{blue!0.0}{\strut ase} \colorbox{blue!0.0}{\strut a} \colorbox{blue!0.0}{\strut gent} \colorbox{blue!21.0}{\strut il} \colorbox{blue!0.0}{\strut icum} \colorbox{blue!0.0}{\strut descent} \colorbox{blue!0.0}{\strut from} \colorbox{blue!0.0}{\strut the} \colorbox{blue!0.0}{\strut down} \colorbox{blue!0.0}{\strut from} \colorbox{blue!0.0}{\strut the} \colorbox{blue!0.0}{\strut ERA} \colorbox{blue!0.0}{\strut ,} \colorbox{blue!0.0}{\strut held} \colorbox{blue!27.0}{\strut the} \colorbox{blue!0.0}{\strut office} \colorbox{blue!41.0}{\strut Carr} \colorbox{blue!0.0}{\strut dime} \colorbox{blue!0.0}{\strut nefarious} \colorbox{blue!0.0}{\strut (} \colorbox{blue!0.0}{\strut or} \colorbox{blue!0.0}{\strut )} \colorbox{blue!0.0}{\strut in} \colorbox{blue!0.0}{\strut Rome} \colorbox{blue!0.0}{\strut around} \colorbox{blue!0.0}{\strut AD} \colorbox{blue!25.0}{\strut 198} \colorbox{blue!0.0}{\strut .} \colorbox{blue!0.0}{\strut listed} \colorbox{blue!0.0}{\strut been} \colorbox{blue!0.0}{\strut a} \colorbox{blue!0.0}{\strut trivial} \colorbox{blue!0.0}{\strut upper} \colorbox{blue!0.0}{\strut Pub} \colorbox{blue!42.0}{\strut l} \colorbox{blue!0.0}{\strut ius} \colorbox{blue!0.0}{\strut Sept} \colorbox{blue!0.0}{\strut im} \colorbox{blue!0.0}{\strut ius} \colorbox{blue!0.0}{\strut Get} \colorbox{blue!0.0}{\strut a} \colorbox{blue!0.0}{\strut ,} \colorbox{blue!0.0}{\strut the} \colorbox{blue!19.0}{\strut brother} \colorbox{blue!0.0}{\strut and} \colorbox{blue!0.0}{\strut rival} \colorbox{blue!22.0}{\strut of} \colorbox{blue!15.0}{\strut emperor} \colorbox{blue!17.0}{\strut .} \colorbox{blue!0.0}{\strut He} \colorbox{blue!0.0}{\strut became} \colorbox{blue!0.0}{\strut one} \colorbox{blue!0.0}{\strut of} \colorbox{blue!0.0}{\strut of} \colorbox{blue!0.0}{\strut earliest} \colorbox{blue!45.0}{\strut Car} \colorbox{blue!0.0}{\strut ac} \colorbox{blue!0.0}{\strut alla} \colorbox{blue!0.0}{\strut 's} \colorbox{blue!37.0}{\strut autical} \colorbox{blue!18.0}{\strut Get} \colorbox{blue!26.0}{\strut a} \colorbox{blue!0.0}{\strut initions} \colorbox{blue!34.0}{\strut Christian} \colorbox{blue!0.0}{\strut has} \colorbox{blue!0.0}{\strut speculated} \colorbox{blue!0.0}{\strut that} \colorbox{blue!0.0}{\strut slight} \colorbox{blue!0.0}{\strut unarmed} \colorbox{blue!32.0}{\strut brightly} \colorbox{blue!0.0}{\strut Pr} \colorbox{blue!0.0}{\strut isc} \colorbox{blue!0.0}{\strut us} \colorbox{blue!0.0}{\strut married} \colorbox{blue!35.0}{\strut possibly} \colorbox{blue!0.0}{\strut sought} \colorbox{blue!0.0}{\strut Caribbean} \colorbox{blue!0.0}{\strut ,} \colorbox{blue!0.0}{\strut a} \colorbox{blue!0.0}{\strut speech} \colorbox{blue!0.0}{\strut close} \colorbox{blue!0.0}{\strut Brom} \colorbox{blue!0.0}{\strut the} \colorbox{blue!0.0}{\strut future} \colorbox{blue!0.0}{\strut .} \colorbox{blue!0.0}{\strut is} \colorbox{blue!0.0}{\strut Pr} \colorbox{blue!0.0}{\strut isc} \colorbox{blue!0.0}{\strut us} \colorbox{blue!36.0}{\strut save} \colorbox{blue!0.0}{\strut a} \colorbox{blue!0.0}{\strut son} \colorbox{blue!0.0}{\strut ,} \colorbox{blue!0.0}{\strut Lucius} \colorbox{blue!0.0}{\strut Mess} \colorbox{blue!10.0}{\strut alla} \colorbox{blue!0.0}{\strut Ap} \colorbox{blue!0.0}{\strut oll} \colorbox{blue!0.0}{\strut in} \colorbox{blue!11.0}{\strut aris} \colorbox{blue!0.0}{\strut ,} \colorbox{blue!0.0}{\strut in} \colorbox{blue!0.0}{\strut ronic} \colorbox{blue!0.0}{\strut .} \colorbox{blue!0.0}{\strut In} \colorbox{blue!0.0}{\strut ge} \colorbox{blue!40.0}{\strut ,} \colorbox{blue!0.0}{\strut and} \colorbox{blue!0.0}{\strut in} \colorbox{blue!31.0}{\strut Roman} \colorbox{blue!14.000000000000002}{\strut Empire} \colorbox{blue!12.0}{\strut ,} \colorbox{blue!0.0}{\strut AD} \colorbox{blue!0.0}{\strut 193} \colorbox{blue!0.0}{\strut -} \colorbox{blue!0.0}{\strut 284} \colorbox{blue!0.0}{\strut (} \colorbox{blue!0.0}{\strut 2011} \colorbox{blue!0.0}{\strut Mess} \colorbox{blue!0.0}{\strut alla} \colorbox{blue!0.0}{\strut Pa} \colorbox{blue!0.0}{\strut etus} \colorbox{blue!0.0}{\strut ,} \colorbox{blue!0.0}{\strut Lucius} \colorbox{blue!0.0}{\strut \#Łobez\#} \colorbox{blue!0.0}{\strut \#Ukrainians\#} \colorbox{blue!6.0}{\strut \#Bernie Sanders\#} \colorbox{blue!0.0}{\strut \#Roman aqueduct\#} \colorbox{blue!0.0}{\strut \#Budapest\#} \colorbox{blue!0.0}{\strut \#Andrés de Santa Cruz\#} \colorbox{blue!48.0}{\strut \textbf{\#Women's Tennis Association\#}} \colorbox{blue!47.0}{\strut \textbf{\#Western text-type\#}} \colorbox{blue!8.0}{\strut \textbf{\#Ukrainians\#}} 
}}}\\ 
    \bottomrule
    \end{tabular}
    \caption{Examples of fusing positions. Top 50 important words are highlighted. The color saturation indicates the importance. Bold and colored words are either pronouns relevant to the topic entity and the phrases wrapped by \# are mentioned entities. }
    \label{tab:case_study_inserted_position_pl}
\end{table*}

\end{document}